\documentclass{sig-alternate}
\usepackage{icvgip}
\usepackage{rotating}

\begin{document}
%
\conferenceinfo{ICVGIP}{'14, December 14-18, 2014, Bangalore, India}
\CopyrightYear{2014} 
\crdata{978-1-4503-3061-5-9/14/12 \$15.00 \\ http://dx.doi.org/10.1145/2683483.2683539}  

\title{ Semantic Motion Segmentation Using Dense CRF Formulation}
%
%
%
%
%

 \icvgipfinalcopy

\def\icvgipPaperID{0208}
\numberofauthors{3}
%
\author{
%
%
\alignauthor
N Dinesh Reddy\\
\affaddr{Robotics Research Center}\\
\affaddr{IIIT Hyderabad, India}\\
\alignauthor
Prateek Singhal \\
\affaddr{Robotics Research Center}\\
\affaddr{IIIT Hyderabad, India}\\
\alignauthor K Madhava Krishna
\affaddr{Robotics Research Center}\\
\affaddr{IIIT Hyderabad, India}\\
}

\maketitle
\graphicspath{{Images/}}

\begin{abstract}
While the literature has been fairly dense in the areas of scene understanding and semantic labeling there have been few works that make use of motion cues to embellish semantic performance and vice versa. In this paper, we address the problem of semantic motion segmentation, and show how semantic and motion priors augments performance. We propose an algorithm that jointly infers the semantic class and motion labels of an object. Integrating semantic, geometric and optical flow based constraints into a dense CRF-model we infer both the object class as well as motion class, for each pixel. We found improvement in performance using a fully connected CRF as compared to a standard clique-based CRFs. For inference, we use a Mean Field approximation based algorithm. Our method outperforms recently proposed motion detection algorithms and also improves the semantic labeling compared to the state-of-the-art Automatic Labeling Environment algorithm on the challenging KITTI dataset especially for object classes such as pedestrians and cars that are critical to an outdoor robotic navigation scenario.
 
\end{abstract}

\section{Introduction}
\label{sec:Intro}
   Using object class and motion cues jointly provides for an enhanced understanding of the scene. We perceive better when we describe the scene in terms of a moving or stationary car (pedestrian) than in terms of presence of only few object classes. Motivated by this fact, we have formulated the problem of object class segmentation, which assigns an object label such as road or building to every pixel in the image, and motion segmentation, in which every pixel within the image is labelled moving or stationary, jointly. Semantic motion segmentation has its application in robotics where an autonomous system will be in a better position to plan its path based on the joint knowledge.

   In this work, we propose a method to model the whole image scene using a fully connected multi-label Conditional random field(CRF) with joint learning and inference. To justify the fact that these problem need to be solved jointly, we find a relation between the layers. As correct labelling of object class can infer the motion labels for the corresponding pixels and motion in the image improves the inference of the object labelling. To provide some intuition behind this statement, note that the object class boundaries are more likely to occur at a sudden transition in motion and vice-versa. Moreover, the class of the object provides a very important clue for motion analysis. For example, in a scene we can assume that the probability of a car or person moving is greater than the probability of a moving wall or a moving road.

   We use sequential stereo pairs from three time instants to label the scene and estimate the motion, showing the robustness in our motion segmentation method. The interaction between the semantic labelling and motion likelihood is learnt, which helps us to efficiently segment the distant moving objects in few time instants. Each image pixel is labelled with both an object class and motion estimate. Various approximate methods for inference exist, such as maximum a posterior methods (e.g graph-cuts), or variational methods, such as mean-field approximation, which allow us to approximately estimate a maximum posterior marginals solution (MPM). We have used mean field based inference algorithm as it enables us to utilize efficient approximations for high-dimensional filtering, which reduce the complexity of message passing from quadratic to linear, resulting in inference that is linear in the number of variables and sub-linear in the number of edges.

   Herein we show that joint labeling formulation is mutually beneficial for motion as well as semantic labeling.  Our method is similar to \cite{bb53700} \cite{DenseObjAtt_CVPR2014} \cite{DBLP:conf/eccv/KimSKS12} where they have used a multi layer multi-label CRF for joint estimation of scene reconstruction and attributes respectively. Specifically we show significant performance gain for motion labeling in the challenging KITTI street datasets in comparison to the state of the art methods in motion segmentation \cite{NamdevKKJ12}. Concurrently we also improve the performance of ALE \cite{LadickyRKT09} by showing segmentation results closer to ground truth especially for pedestrians and cars. We accomplish this using motion likelihood estimates and incorporate semantics to get a better holistic understanding of the dynamic scene. These results show that this approach closely mimics perception by humans where semantic labels play an important role to identify motion.

\begin{figure*}[t!]
\centering
\includegraphics[width=160mm]{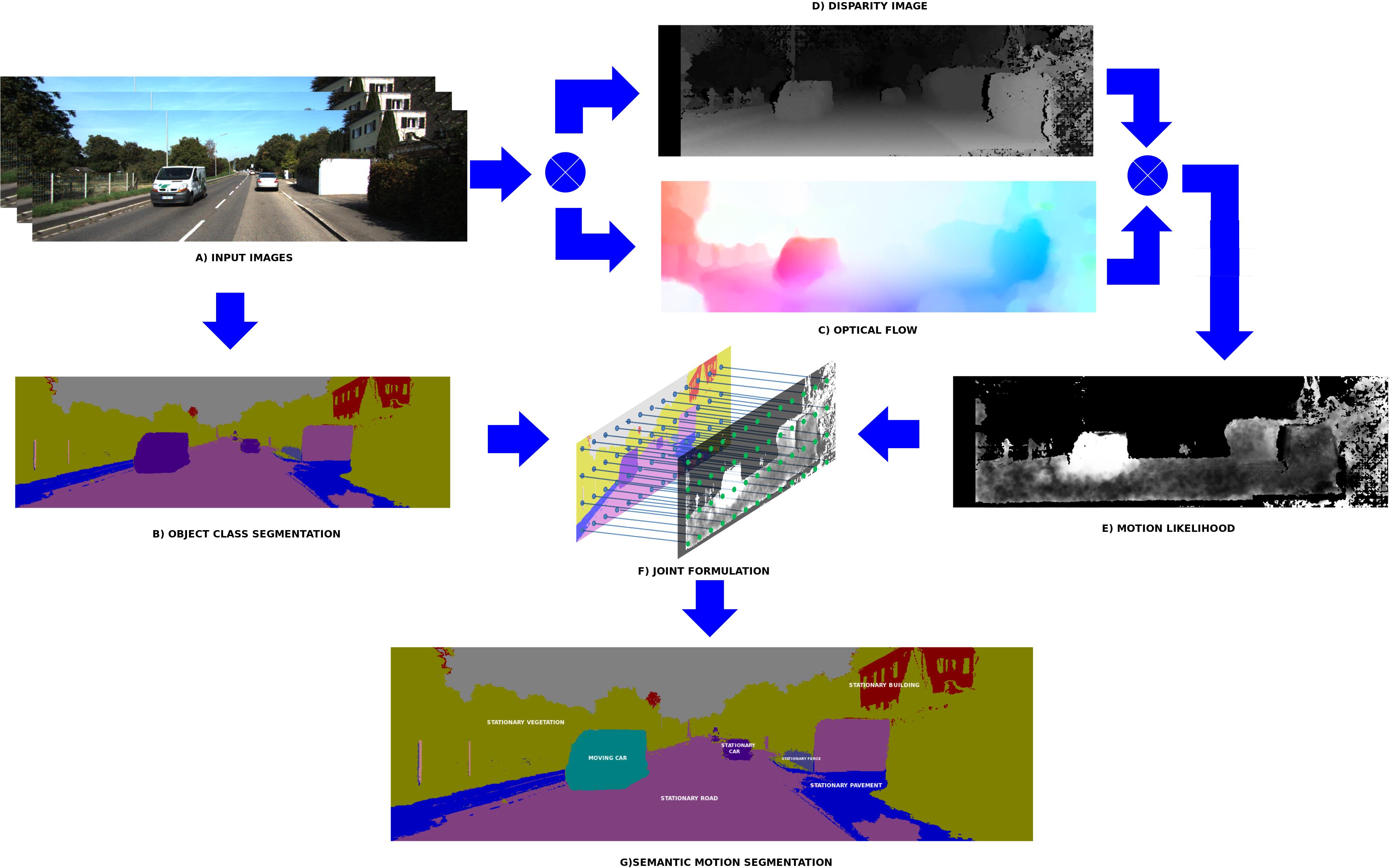}
\caption{
Illustration of the proposed method .The system takes a sequence of rectified stereo images from the tracking dataset of KITTI (A).Our formulation computes the Object class probabilities (B) and motion likelihood (E) using disparity map(D) and optical flow(C).These are input into a joint formulation which exploits the object class and motion co-dependencies by allowing a interact between them (F).The inference is computed using the mean field approximation method to give a joint label to each pixel(G).  Best viewed in color.
}
\label{fig:pipeline}
\end{figure*}

\section{Related Work}

There has been fairly large amount of literature in both Semantic and Motion segmentation. For semantic image segmentation, existing approaches use textonboost \cite{Shotton06textonboost:joint}, in which weakly predictive features in a image like color, location and texton features are passed through a classifier to give the cost of a label for that particular pixel. These costs are combined in a contrast sensitive Conditional Random field \cite{lafferty01conditional} \cite{Koller:2009:PGM:1795555}. Most of the mid-level inferences do not use pixels directly, but segment the image into regions \cite{KohliKT09}\cite{KohliLT08} \cite{conf/iccv/GouldFK09} \cite{LadickyRKT09}. Substantial state of the art results for dense semantic image segmentation have been show using superpixel based hierarchical framework \cite{LadickyRKT10}\cite{bb53700}. Recently a lot of the scene understanding research has gone into better understanding the scene using different parameters to get a better segmentation. In \cite{Ladicky:2010:MCO:1888089.1888122},\cite{LadickyRKT10},\cite{bb53700} Ladicky et al have used Object detectors, Co-occurrence statistics and Stereo disparity for improving the semantics. In \cite{YaoFU12}, Yao et al have combined semantic segmentation, object detection and scene classification for understanding a scene as a whole .

   Motion segmentation has been approached using geometric priors mostly from a video. General paradigm involves using Geometric constraints \cite{NamdevKKJ12} or reducing the model to affine to cluster the trajectories into subspaces  \cite{Elhamifar09sparsesubspace}. These methods have been shown not to work in complex environments where the moving cars lie in the same subspace. We consider deviation of the trajectories based on the 3d motion of the camera estimated from the trajectories to provide us motion likelihood even in challenging scenarios. Semantics for motion detection is not new, Wedel et al  \cite{DBLP:conf/emmcvpr/WedelMRFC09} have used scene flow to segment motion in a stereo camera. There has been numerous work on segmenting moving object by compensating for the platform movement \cite{NamdevKKJ12} \cite{Lenz2011IV}\cite{Romero-CanoN13}.Recently in \cite{Icra2014deep} motion features have been learnt using deep learning to give better motion likelihood estimate. The applications for understanding motion semantics has been an emerging area and can be used to understand and model traffic pattern\cite{Geiger2012CVPR} \cite{Zhang2013ICCV} \cite{Geiger2014PAMI}.
       
    The structure of the paper is as follows: Section {\ref{sec:CRF}} formulates the CRF's for dense image labelling, and describe how they can be applied to the the problem of object class segmentation and motion segmentation. Section {\ref{sec:joint_crf}} describes the joint formulation allowing for the joint optimization of these problems, while Section {\ref{sec:inference}} describes the mean field inference for the joint optimization. The learning for the class and motion correlation is described in Section {\ref{sec:learning}}. We evaluate and compare our algorithm in Section {\ref{sec:evaluation}}.

\section{Multi-Label CRF formulation}
\label{sec:CRF}
Our joint optimisation consists of two parts, object class segmentation and motion segmentation. We introduce the terms to be used in the paper. We define a dense CRF where the set of random variables $Z=\lbrace Z_1,Z_2,....,Z_N \rbrace$  
corresponds to the set of all image pixels $i \in \mathcal{V} = \lbrace 1,2,...,N\rbrace$ . Let $\mathcal{N} $
be the neighbourhood system of the random field defined by the sets $\mathcal{N}_i \forall i \in \mathcal{V} $
, where $\mathcal{N}_i $ denotes the neighbours of the variable $Z_i$ . Any possible assignment of labels to the random variables will be called a labelling and denoted by $z$.

\subsection{Dense Multi-class CRF}

We formulate the problem of object class segmentation as finding a minimal cost labelling of a CRF defined over a set of random variables $X=\lbrace x_i,x_2,....,x_\mathcal{N}\rbrace$ each taking a state from the label space $\mathcal{L}=\lbrace l_1,l_2,....,l_k\rbrace$, where k represents the number of object class labels.
 Each label $l$ indicates a different object class such as car, road ,building or sky.These energies are:
\begin{align}
E^O  (x)=\sum_{i \in \mathcal{V}} \psi^O_i (x_i) +\sum_{i \in \mathcal{V} , j \in \mathcal{N}_i} \psi^O_{i,j} (x_i,x_j)
\end{align}

The unary potential $\psi^O_i (x_i)$ describes the cost of the pixel taking the corresponding label. The pairwise potential encourages similar pixels to have the same label. The unary potential term is computed for each pixel using pre-trained models of the color, texture and location features for each object \cite{Shotton06textonboost:joint} .In a typical graph topology, we consider a 4 or 8 neighbour connected network. With the mean field inference algorithm it is possible to use a fully connected graph, where all the pixels in the image are interconnected given certain forms of pairwise potential.Therefore, the pairwise potential takes the form of a potts model:
\begin{align}
  \psi^O_{i,j} (x_i,x_j) =\left\{ \begin{array}{ll}
         0 & \mbox{if ${x_i=x_j}$};\\
        p(i,j) &  \mbox{if $x_i \neq x_j$}.\end{array} \right. \
\end{align}

For a fully connected graph topology, $p(i,j)$ is given as:
\begin{align}
  p(i,j)= exp(-\frac{|p_i-p_j|^2}{2 \theta_{\beta}^2} -\frac{ I_i-I_j}{2\theta_v^2})  + exp(-\frac{|p_i-p_j|^2}{2 \theta_p^2})
\end{align}
Where $p_i$ indicates the location of the ith pixel, $I_i$ indicates the intensity of the $i_{th}$ pixel, and $\theta_{\beta}$,$\theta_p$ ,$\theta_v$ are the model parameters learned from the training data.

\subsection{Dense Motion CRF}

We use a standard dense CRF for formulating the semantic motion segmentation .The problem is posed as finding a minimal cost labelling of a CRF over a set of random variables $\mathcal{Y}=\lbrace y_1 ,y_2,....,y_\mathcal{N} \rbrace$ which can take the label of moving or stationary i.e $\mathcal{M}=\lbrace m_1 , m_2 \rbrace$ . where $m_1$ represents all the stationary pixels in the image and $m_2$ corresponds to the moving pixels. The formulation for motion is as follows:

\begin{align}
E^\mathcal{M}  (y)=\sum_{i \in \mathcal{V}} \psi^\mathcal{M}_i (y_i) +\sum_{i \in \mathcal{V} , j \in \mathcal{N}_i} \psi^\mathcal{M}_{ij} (y_i,y_j)
\end{align}

Where the unary potential $\psi^\mathcal{M}_i (x_i)$ is given by the motion likelihood of the pixel and is computed as the difference between the predicted flow and optical flow. The predicted flow is given by  :

\begin{align}
   \hat{X}'=KRK'X+KT/z
     \label{eq:motion_estimate}
\end{align}

where K is given as the Intrinsic camera matrix , R and T are the translation and rotation of the camera respectively, z is the depth of the pixel from camera. X is the location of the pixel in image coordinates and $\hat{X}'$ is the predicted flow vector of the pixel given from the motion of the camera. Thus unary potential is given as:

\begin{align}
  \psi^\mathcal{M}_i (x_i) =({ (\hat{X}'-X')^T \Sigma^{-1}(\hat{X}'-X') })
  \label{eq:motion}
\end{align}

Where $\Sigma$ is called the covariance matrix which is the sum of covariance of optical flow and the covariance of measured optical flow. Here $\hat{X}'-X'$ represents the difference of the predicted flow and optical flow. The pairwise potential $\psi^\mathcal{M}_{i,j} (y_i,y_j) $ is given as the relationship between neighbouring pixels and encourages the adjacent pixels in the image to have similar flow. The cost of the function is defined as:
\begin{align}
  \psi^\mathcal{M}_{ij} (y_i,y_j) =\left\{ \begin{array}{ll}
         0 & \mbox{if ${y_i=y_j}$};\\
        g(i,j) &  \mbox{if $y_i \neq y_j$}.\end{array} \right. \
\end{align}
where $g(i,j)$ is a edge feature based on the difference between the flow of the neighbouring pixels:
\begin{align}
g(i,j) = |f(y_i)-f(y_j)|
\end{align}
where f(.) is defined as the function which returns the flow vector of the corresponding pixel.
 
\section{Joint CRF Formulation}
\label{sec:joint_crf}
In this section, we try to use object class segmentation and motion estimate to jointly estimate the label of the dynamic scene . Each random variable $Z_i$ = [$X_i$,$Y_i$] takes a label $z_i$ = [$x_i$,$y_i$], from the product space of object class and motion labels and correspond to the variable $Z_i$ taking a object label $x_i$ and motion $y_i$. In general the energy of the CRF for joint estimation is written as :

\begin{align}
E^\mathcal{J}  (z)=\sum_{i \in \mathcal{V}} \psi^\mathcal{J}_i (z_i) +\sum_{i \in \mathcal{V} , j \in \mathcal{N}_i} \psi^\mathcal{J}_{i,j} (z_i,z_j)
\end{align}

where $\psi^\mathcal{J}_i$ , $\psi^\mathcal{J}_{i,j}$ are the sum of the previously mentioned terms  $\psi^O_i$ and  $\psi^\mathcal{M} _i$, $\psi^O_{i,j}$ and  $\psi^\mathcal{M} _{i,j}$ respectively. we include some extra terms which help in understanding the relation between the labels of X , Y . In the real world scenarios there is a relationship between the object class and corresponding motion likelihood for each pixel. we compute an interactive unary and pairwise potential terms so that a joint inference can be performed.

\subsection{Joint unary potential}

The unary potential $\psi_i^\mathcal{J}(z_i)$ can be defined as an interactive potential term which incorporates a relationship between the object class and the corresponding motion likelihood. we can directly take the relationship between the object class and all the possible motion models as a measure to calculate the joint unary potential. As this requires large amount of training data to incorporate all motion models for each the class. we look at class and motion correlation function which incorporates the class-motion compatibility and can be expressed as :
\begin{align}
  \psi^{O \mathcal{M}}_{i,l,m} (x_i,y_j) =    \lambda(l,m);
    \label{eq:unary_joint}
\end{align}

Here $\lambda(l,m) \in [-1,1]$ is a learnt correlation term between the motion and object class label.The combined unary potential of the joint CRF is given as follows:

\begin{align}
\psi_{i,l,m}^\mathcal{J}([x_i,y_i])= \psi_i^O(x_i)+ \psi_i^\mathcal{M}(y_i) + \psi_{i,l,m}^{O \mathcal{M}}(x_i,y_i)
\end{align}

where $\psi^O_i$ and $\psi_i^\mathcal{M}$ , are the unary potentials previously discussed for object class and motion likelihood of a pixel $i$ given the image.



\subsection{Joint pairwise potential}

The joint pairwise potential $\psi_{ij}^\mathcal{J}(m_i,m_j)$ enforces the consistency of object class and motion between the neighbouring pixels. This potential term exploits the condition that ,when there is a change in the motion layer , then there is a high chance for the label of the object class to change .Similarly if there is a change in the label in the object class then it is more likely for the label in the motion layer to change .To include this behaviour in our formulation we have taken the joint pairwise term as:
\begin{align}
\psi^\mathcal{J}_{ij}([x_i,y_i],[x_j,y_j]) =
 \psi_{ij}^O(x_i,x_j)
 + \psi_{ij}^\mathcal{M}(y_i,y_j)
\end{align}
Here $\psi_{ij}^O(x_i,x_j)$ and $\psi_{ij}^\mathcal{M}(y_i,y_j)$ have been defined earlier as the pairwise terms of object class and motion respectively.

\section{Inference}
\label{sec:inference}
The inference has been a challenging problem for large scale CRFs. We follow Krahenbuhl et al {\cite{Krahenbuhl_Koltun_2011}} {\cite{conf/icml/KraehenbuehlK13}},which uses a mean field approximation approach for inference . The mean-field approximation introduces an alternative distribution over the random variables of the CRF,$Q_i (z_i)$ , where the marginals are forced to be independent $Q(z)=\prod_i Q_i (z_i)$. The mean-field approximation then attempts to minimize the KL-divergence between the Q and the true distribution P. We can therefore take $Q_i (z_i)=Q^O_i(x_i) Q_i^\mathcal{M}(y_i)$ . Here $Q^O_i$ is a multi-class distribution over the object labels , and $Q_i^\mathcal{M}$ is a binary distribution over moving or stationary given by $\lbrace 0,1\rbrace$.

\begin{align*}
Q_i^O(x_i=l) \  = \ & 1/Z_i \ \exp\lbrace - \psi_i^O(x_i) \\
                &   - \sum_{l' \in \mathcal{L}} \sum_{i \neq j} Q_i^O(x_j = l').\psi_{ij}^O(x_i,x_j)\\
                & - \sum_{m' \in \mathcal{M}} Q_i^\mathcal{M}(y_i = m').\psi_{i,l,m}^{O\mathcal{M}}(x_i,y_i)\rbrace
\end{align*}

The inference for the Motion layer is similar to the object class layer and is given by:

\begin{align*}
Q_i^\mathcal{M}(y_i=m) \  = \ & 1/Z_i \ \exp\lbrace - \psi_i^\mathcal{M}(y_i) \\
                &   - \sum_{m' \in \mathcal{M}} \sum_{i \neq j} Q_j^\mathcal{M}(y_j = l') \psi_{ij}^\mathcal{M}(y_i,y_j)\\
                & - \sum_{l' \in \mathcal{L}} Q_i^O(x_i = m').\psi_{i,l,m}^{O\mathcal{M}}(x_i,y_i)\rbrace
\end{align*}

Where $Z_i$ is given as the normalization factor , and $m \in \lbrace 0,1 \rbrace$. As proposed in {\cite{Krahenbuhl_Koltun_2011}}, Using $n+m$ Gaussian convolutions we can efficiently evaluate the pairwise summations which are given as Potts model.



\section{Learning}
\label{sec:learning}
We learn the parameters for the label and motion in this section. We describe a piecewise method for training the label and motion correlation matrices. In the  model described, we train for the matrix simultaneously by learning an $(n+2)^2$ correlation matrix.\\
     We use the modified adaboost framework implemented in \cite{DenseObjAtt_CVPR2014}. For training we denote the training dataset of $N$ instances of pixels or regions as $\mathcal{D} = \lbrace (\bold{t_1},\bar{z}_1),(\bold{t_2},\bar{z}_2),...,\\(\bold{t}_\mathcal{N},\bar{z}_\mathcal{N}))$. Here, $t_i$ is a feature vector for the $i$-th instance  and $\bar{z_i}=[\bar{x_i},\bar{y_i}]$ is an indicator vector of length $n+2$ , where $\bar{x_i}(l)=1$ implies that the class label is associated with the pixel or region instance of $i$ and $\bar{x_i}(l)=-1$ represents that the class is not associated with the instance $i$ and similarly for $\bar{y_i}(m)=1$ and $\bar{y_i}(m)=-1$ represents the association of motion $m$ for the instance $i$. Therefore, $\bar{z_i}$ represents the object class and motion ground truth information for the instance $i$.
     
    In the following approach, we show how to compute $\lambda(l,m)$ .The boosting approach in \cite{Huang:2012:MHR:2339530.2339615} generates a strong classifier $H_{s,l}(t)$ for each object class $l$ and each round of boosting $s=1,2,3,....,S$.These strong classifiers can be defined as:

\begin{align*}
H_{s,l}(t)=\sum_{s=1,2,...,S} \alpha_{s,l}h_{s,l}(\bold{t})
\end{align*}      
     
     Here $h_{s,l}$ are weak classifiers , and $\alpha_{s,l}$ are the non-negative weights set by the boosting algorithm.As proposed in \cite{Huang:2012:MHR:2339530.2339615} , we use their joint learning approach, which generates a sequence of reuse weights $\beta_{s,l}(H_{s-1},m)$ for each class and motion attributes $l,m$ at each iteration $s$. These represent the weight given to the strong classifier for motion label m in round $s-1$ in the classifier for $l$ at round $s$. Using the following reuse weights and the strong classifiers we can calculate the label correlation :
\begin{align*}
\lambda (l,m)=\sum_{s=2,...S} \alpha_{s,l}(\beta_{s,l}(H_{s-1,m})) - \beta_{s,l}(-H_{s-1,m}))
\end{align*}      
     This learning approach incorporates information about the motion likelihood and appearance relationship between motion and objects.

\begin{figure*}[t!]
\begin{center}
\begin{tabular}{c c c c}
\begin{sideways}\bf \centering  \ INPUT\end{sideways} & \includegraphics[width=50mm]{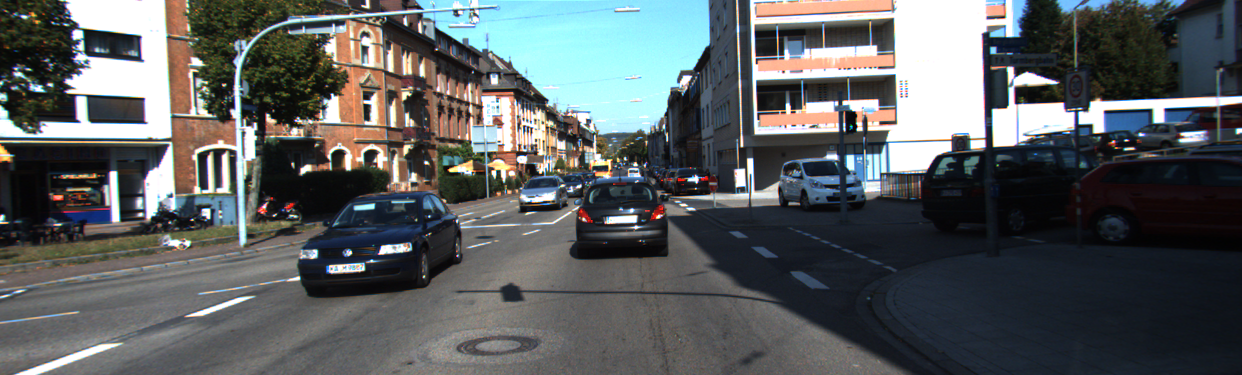} & \includegraphics[width=50mm]{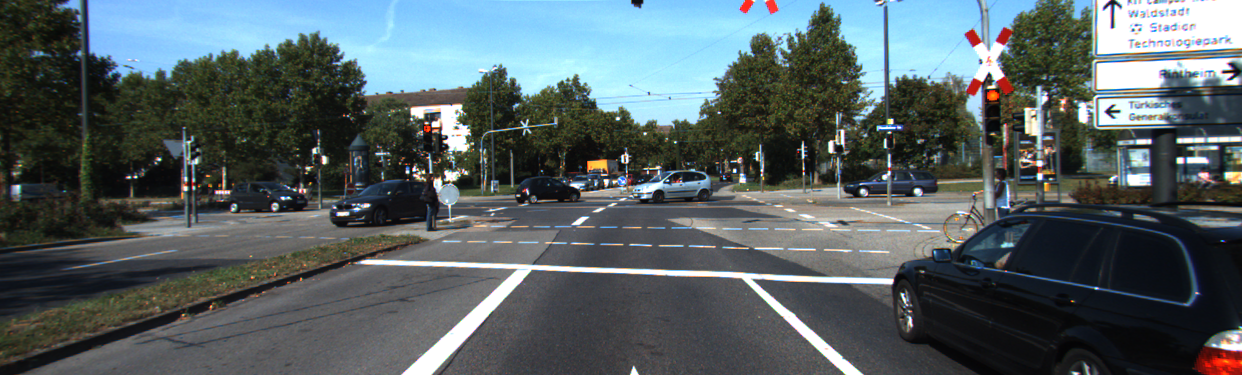} & \includegraphics[width=50mm]{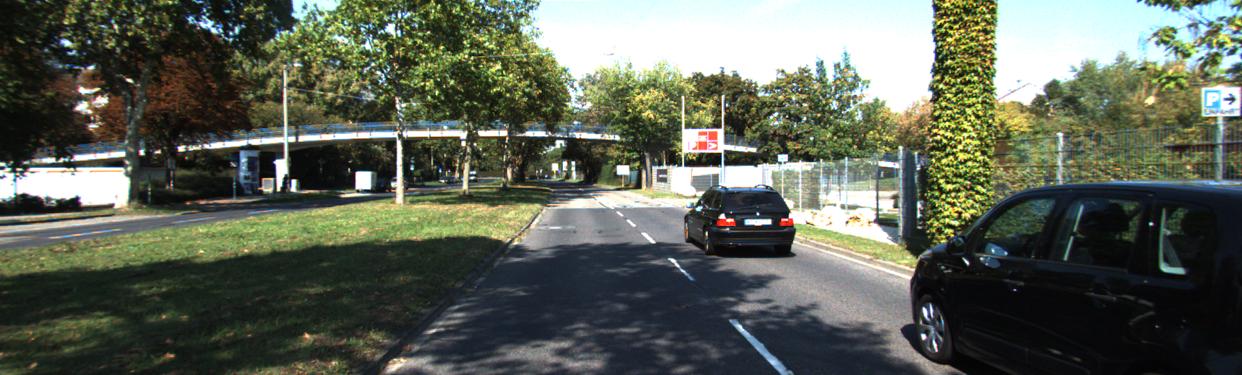}  \\
\begin{sideways}\bf \centering   \ GT-O\end{sideways} & \includegraphics[width=50mm]{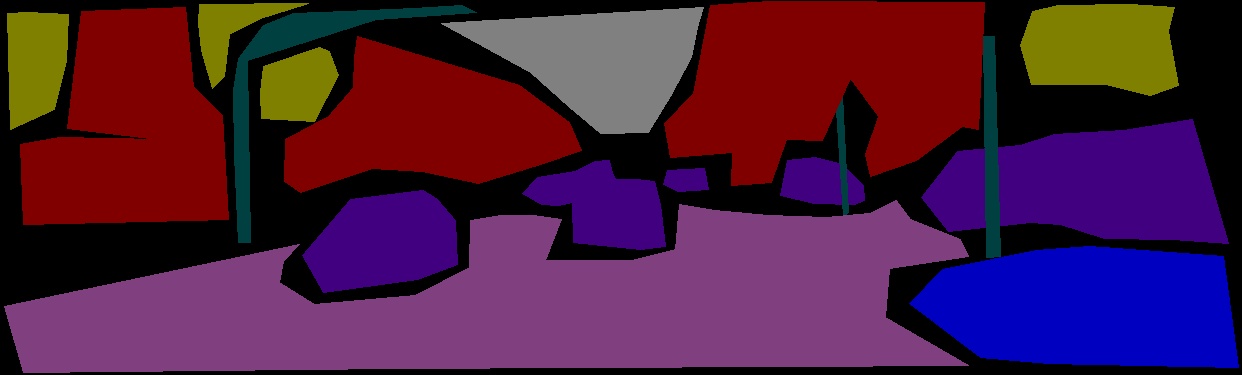} & \includegraphics[width=50mm]{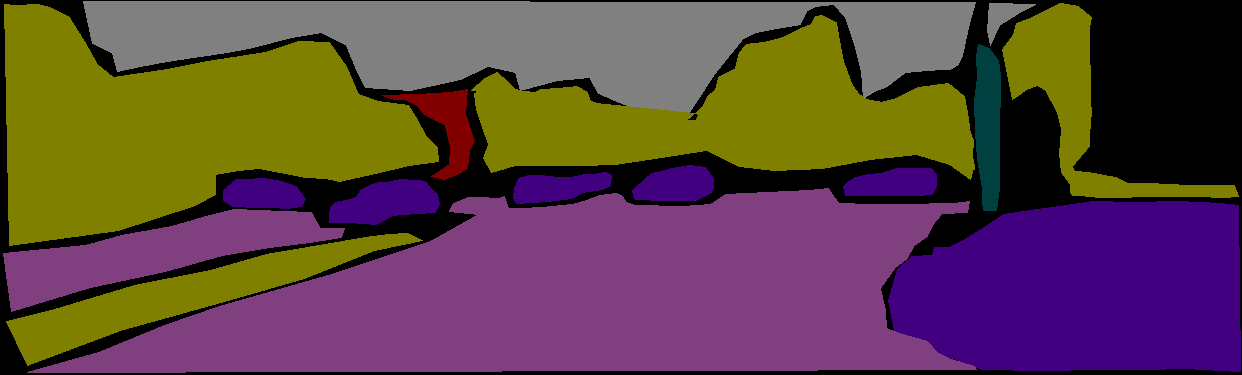} & \includegraphics[width=50mm]{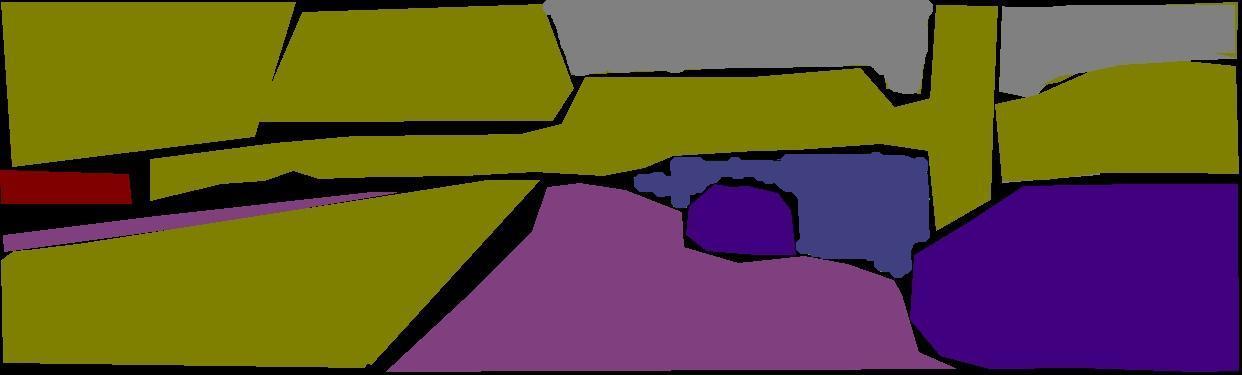}\\
\begin{sideways}\bf\centering \ FULL-C\end{sideways} & \includegraphics[width=50mm]{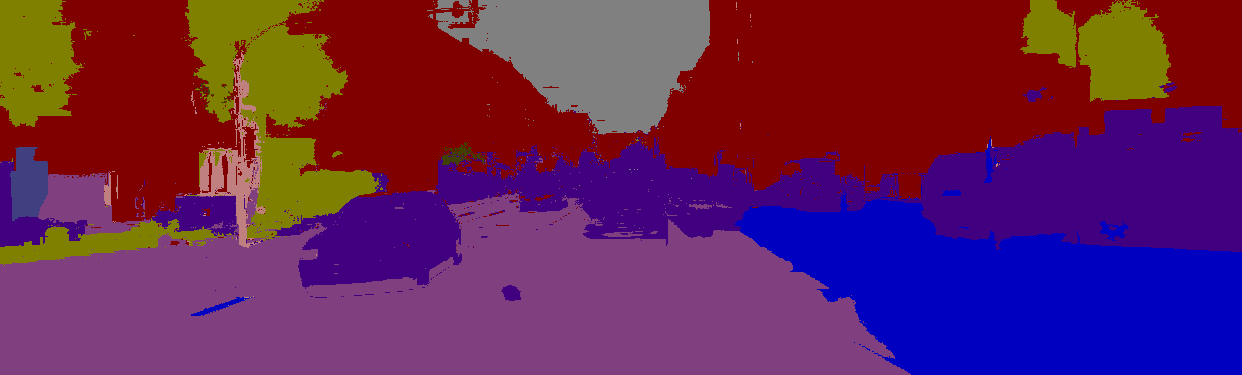} & \includegraphics[width=50mm]{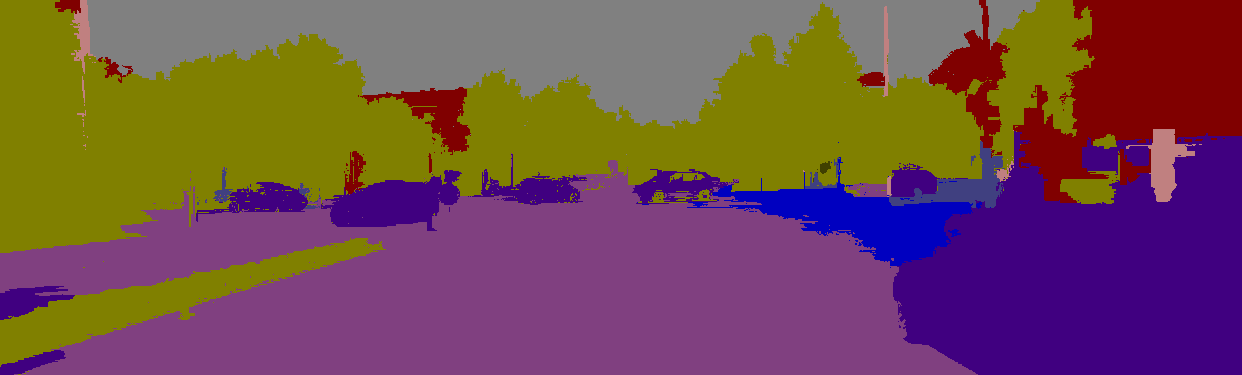} & \includegraphics[width=50mm]{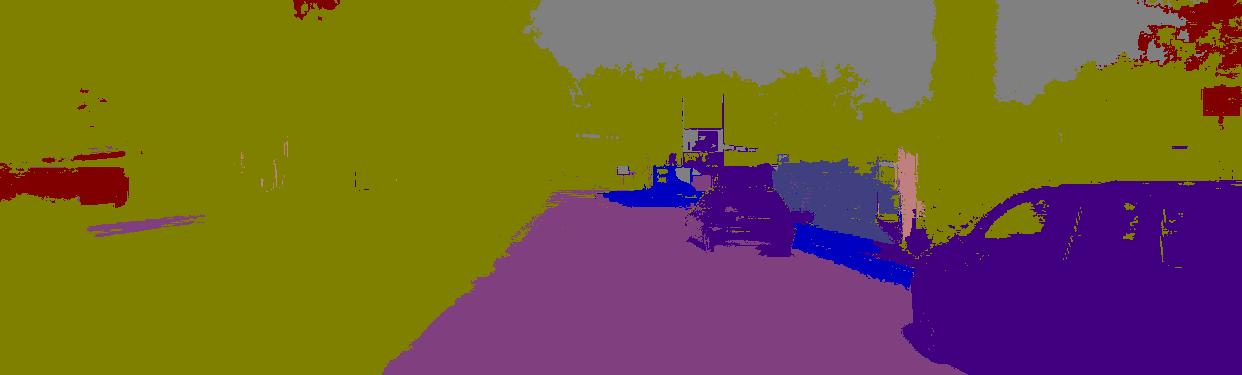}\\
\begin{sideways}\bf    OURS-O\end{sideways} & \includegraphics[width=50mm]{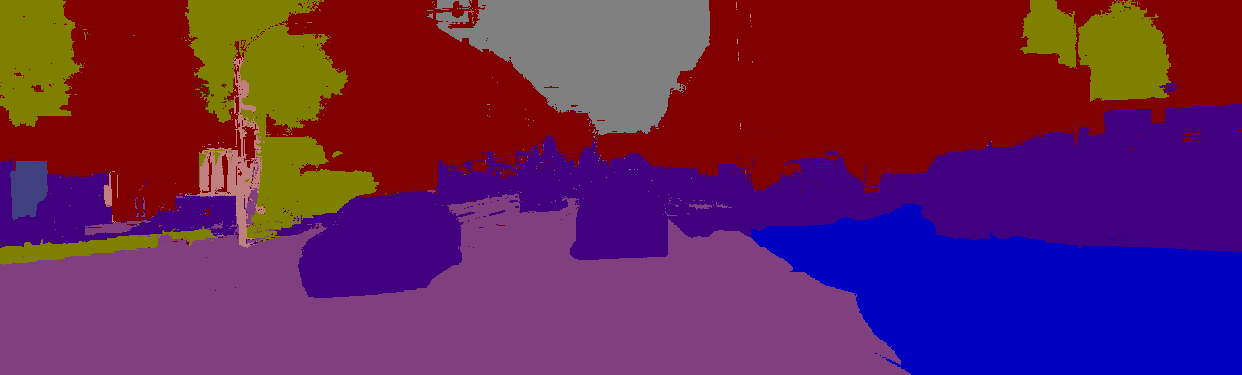} & \includegraphics[width=50mm]{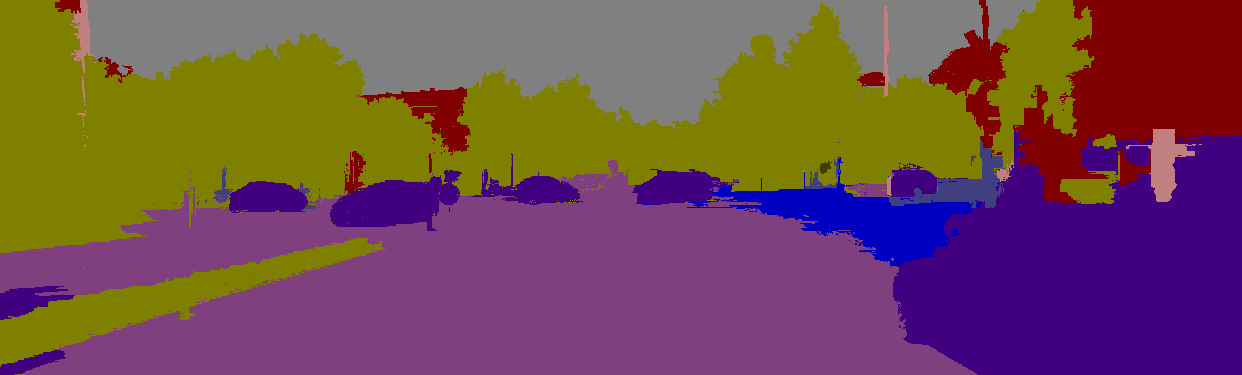} & \includegraphics[width=50mm]{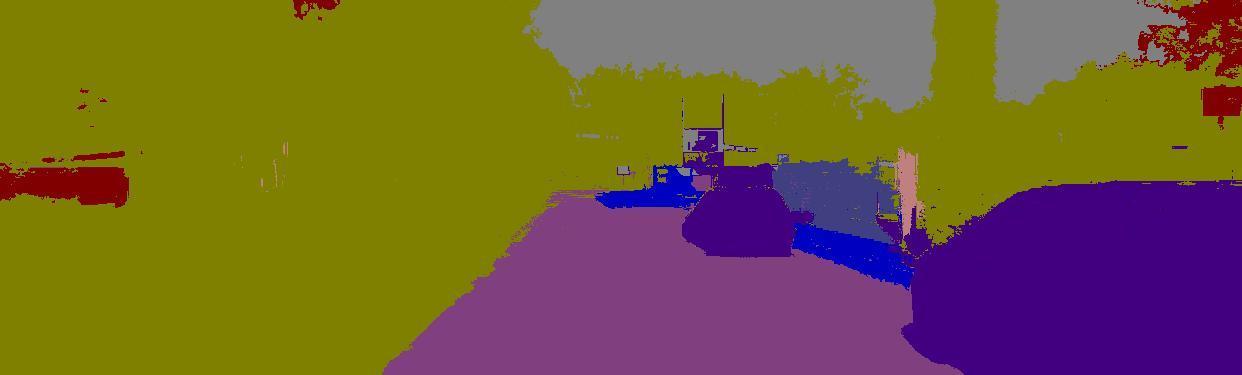} \\
\begin{sideways}\bf  \ \ GT-M\end{sideways} & \includegraphics[width=50mm]{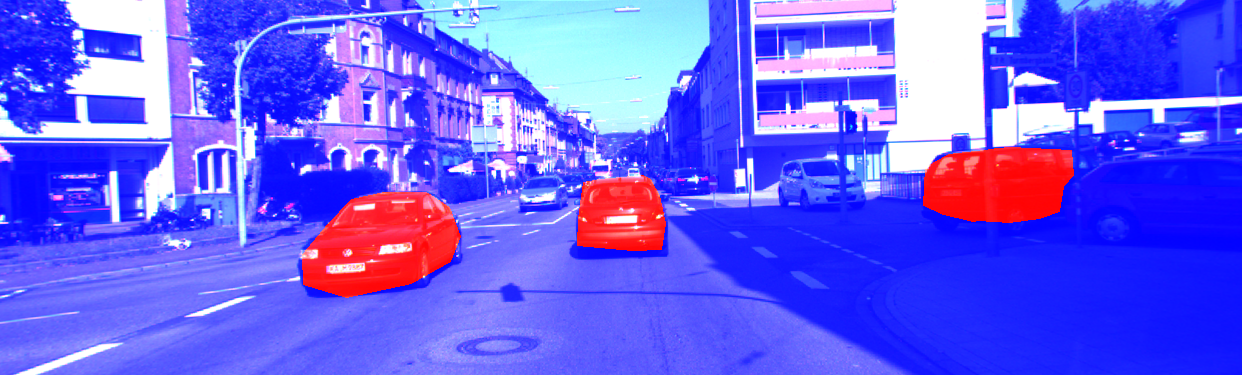} & \includegraphics[width=50mm]{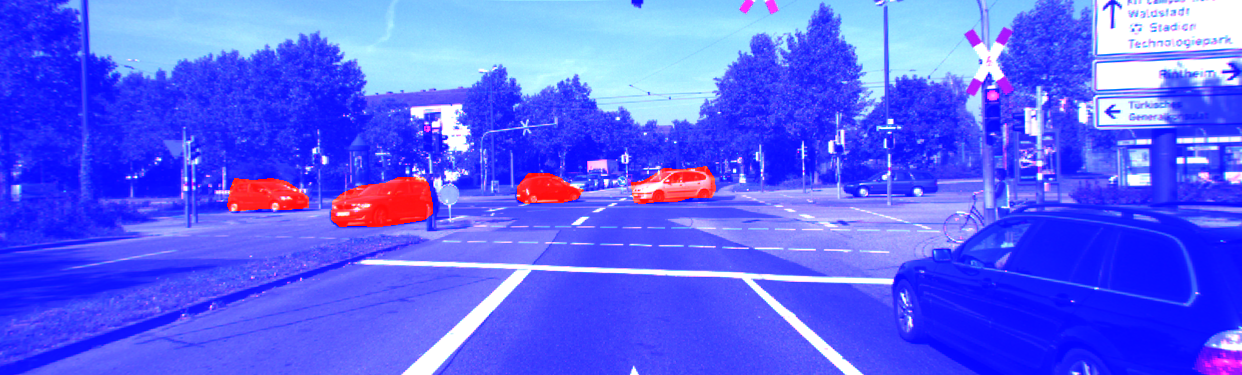} & \includegraphics[width=50mm]{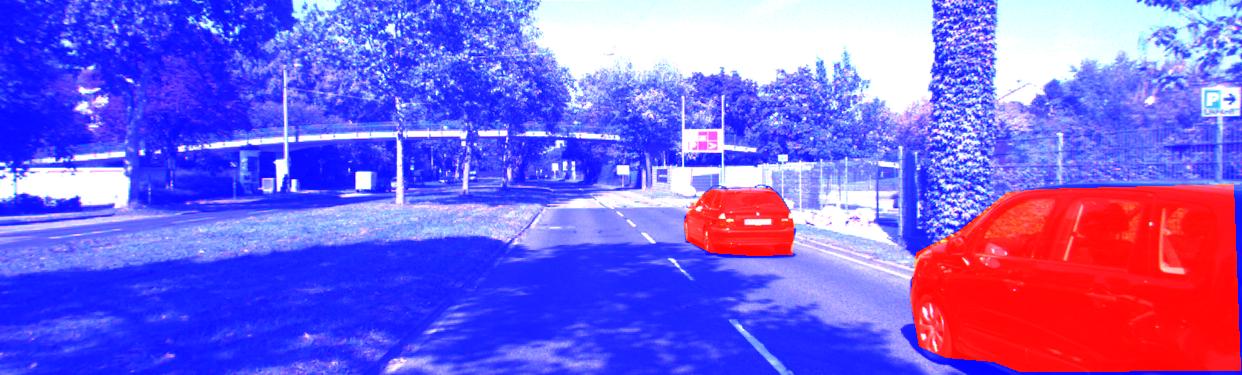}\\
\begin{sideways}\bf  \  GEO-M \end{sideways} & \includegraphics[width=50mm]{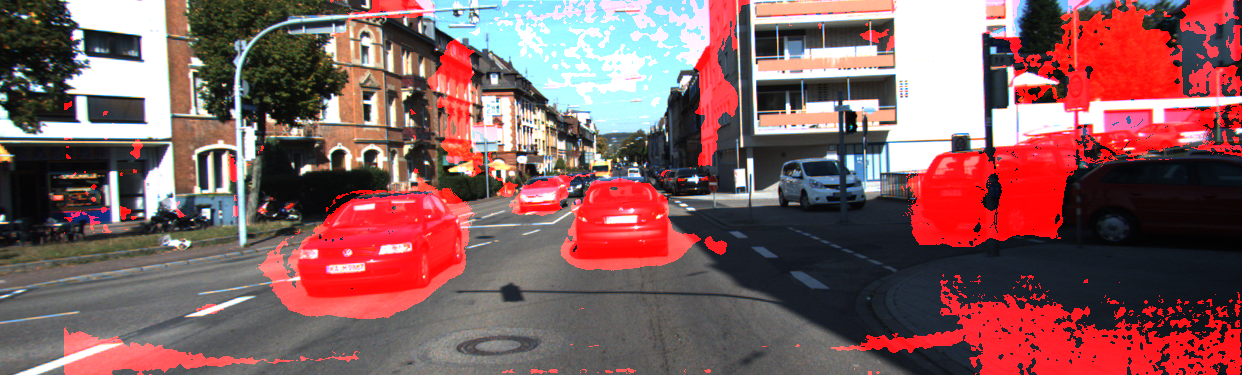} & \includegraphics[width=50mm]{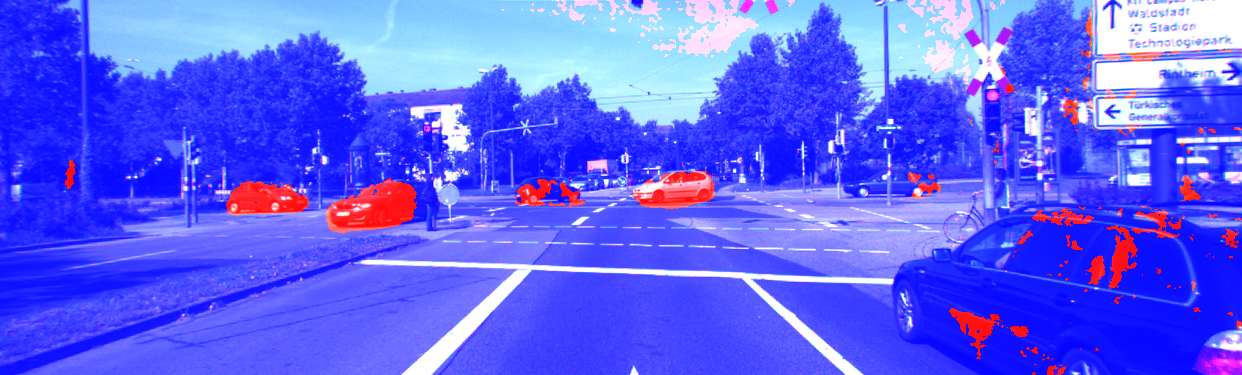} & \includegraphics[width=50mm]{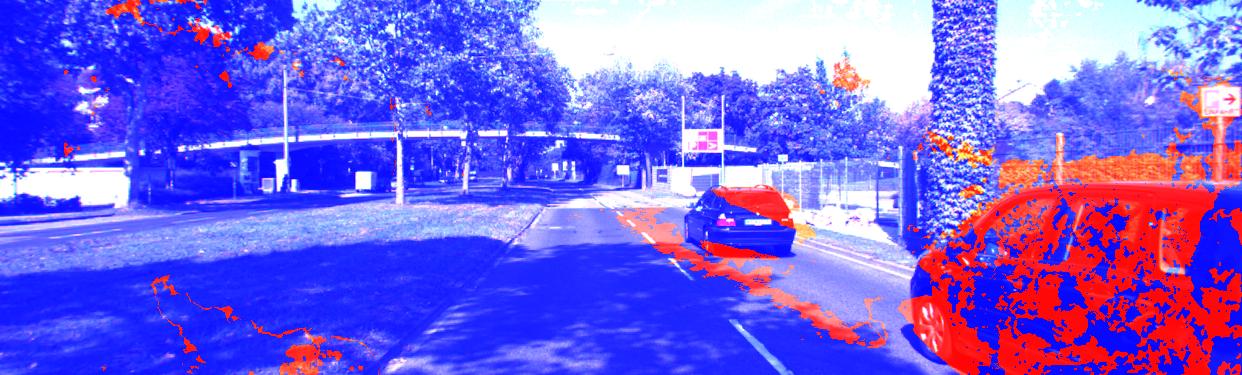}\\
\begin{sideways}\bf   OURS-M\end{sideways} & \includegraphics[width=50mm]{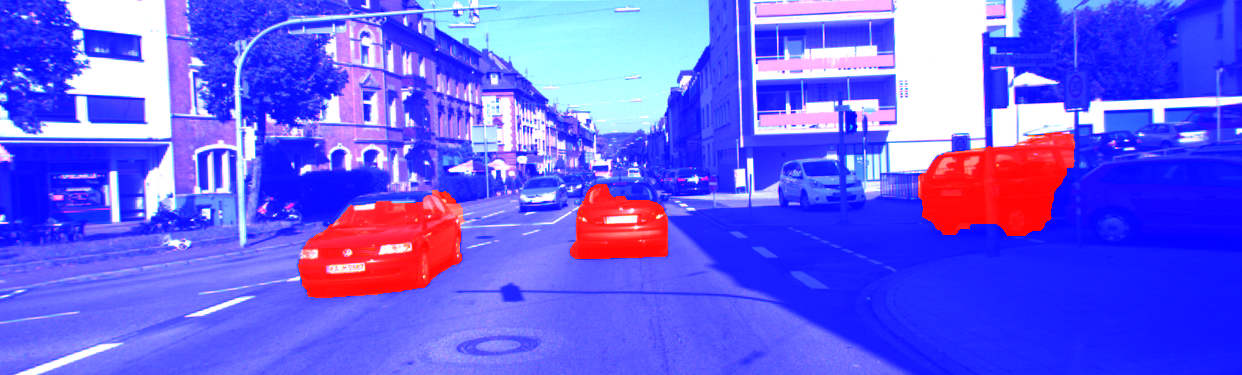} & \includegraphics[width=50mm]{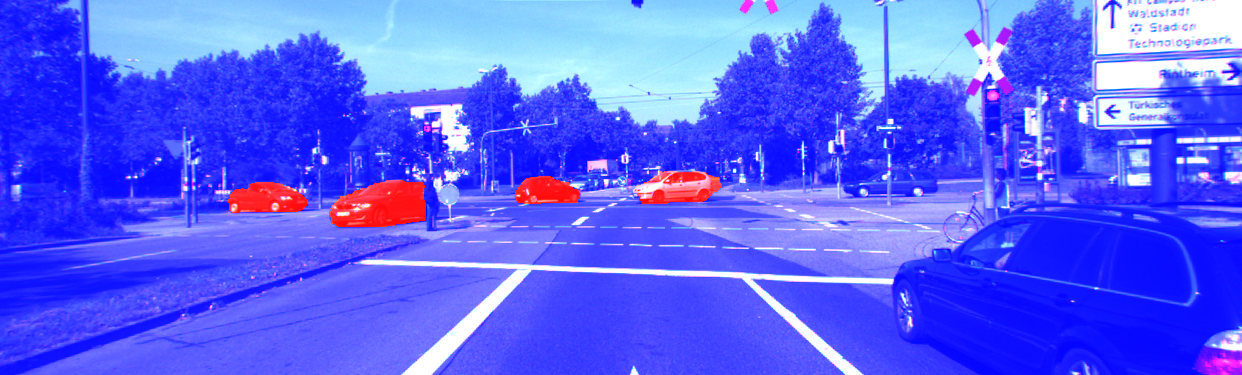} & \includegraphics[width=50mm]{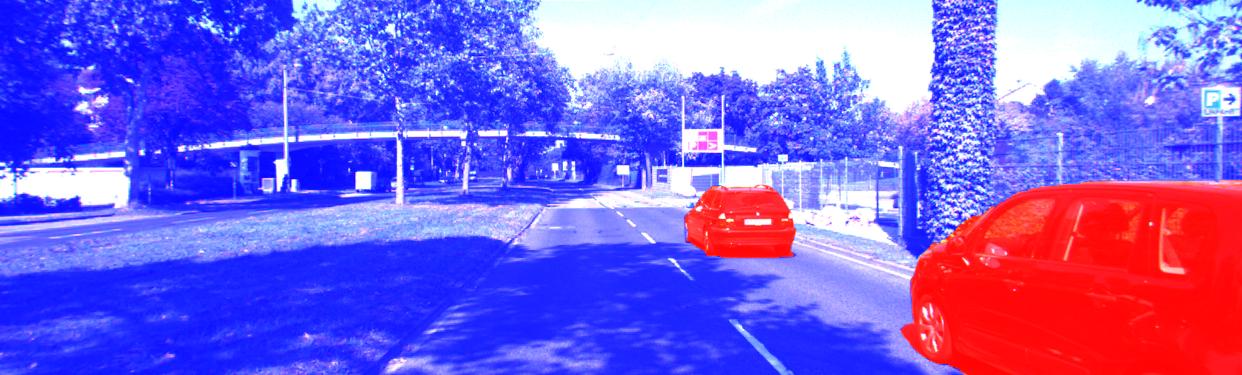}\\
& \multicolumn{3}{c}{\includegraphics[width=158mm]{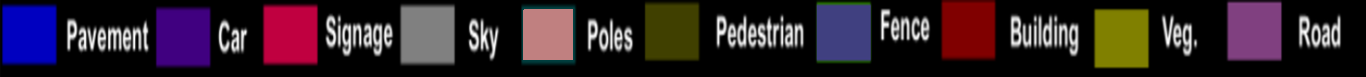}}\\
 & \bf Sequence 1 & \bf Sequence 2 & \bf Sequence 3  \\
\end{tabular}

\end{center}
\label{fig:ale_check123}

 \caption{
Qualitative object class and motion results for the KITTI dataset 1) Images of three sequences of KITTI dataset(INPUT) 2)Ground truth of object class segmentation (GT-O) 3)Object class segmentation results using fully connected CRF(FULL-C) 4)Object class segmentation using the Joint formulation of the proposed method (OURS-O) 5) Ground truth of the motion segmentation (GT-M) 6)Motion segmentation using geometric constraints (GEO-M)\cite{NamdevKKJ12} 7) Proposed method dense motion segmentation(OURS-M).For motion segmentation blue depicts stationary and red pixels represent moving. best viewed in color
}
\end{figure*}

\section{Experimental Evaluation}
\label{sec:evaluation}
We have used a popular street-level dataset---\textit{KITTI} for evaluation. It consists of several sequences collected by a car-mounted camera driving in urban,residential and highway environments, making it a varied and challenging real world dataset. We have taken 6 sequences from the tracking dataset of KITTI each containing 20 stereo images, each of size 1024 x 365. Firstly, these sequences were manually annotated with the 11 object classes containing the spectrum of classes. Secondly, each of the image was annotated with moving and non-moving objects using the tracking ground truth data. These sequences are challenging as they contain multiple moving cars and the labels consisted of 11 classes \textit{ i.e Pavement, Car, Signal, Sky, Poles, Pedestrian, Fence, Building, Vegetation and Road}. We have selected KITTI dataset as it contains stereo image pairs with a wide baseline.We Learned the motion compatibility, as simple lookup table would not work due to instances where the semantic prior is wrong. The hard negatives provide us with the ability to categorically remove objects with wrong motion likelihood a common occurrence due to inconsistent disparity. This would allow us to test on a variety of datasets without needing to train for similar classes . The dataset was also chosen with the view to showcase the algorithm’s capability on degenerate cases which are not commonly addressed in other datasets.

  We have used semi-global block matching \cite{Hirschmueller08} disparity map computation algorithm for the disparity computation in the stereo camera sequence. For the computation of the motion of the moving camera, we have used RANSAC based algorithm to solve for the Eq.\eqref{eq:motion_estimate}. We have added the temporal consistency of motion across 3 images to the likelihood estimate which improves the results. As for the dense optical flow computation in the implementation, we have used the Deepflow algorithm from {\cite{weinzaepfel:hal-00873592}}, which has given state-of-the art results for the KITTI evaluation benchmark. For object class segmentation we have used the publicly available Texton boost classifier to compute the unary potentials for each class.

\begin{figure}[!t]
\begin{center}
\includegraphics[width=80mm]{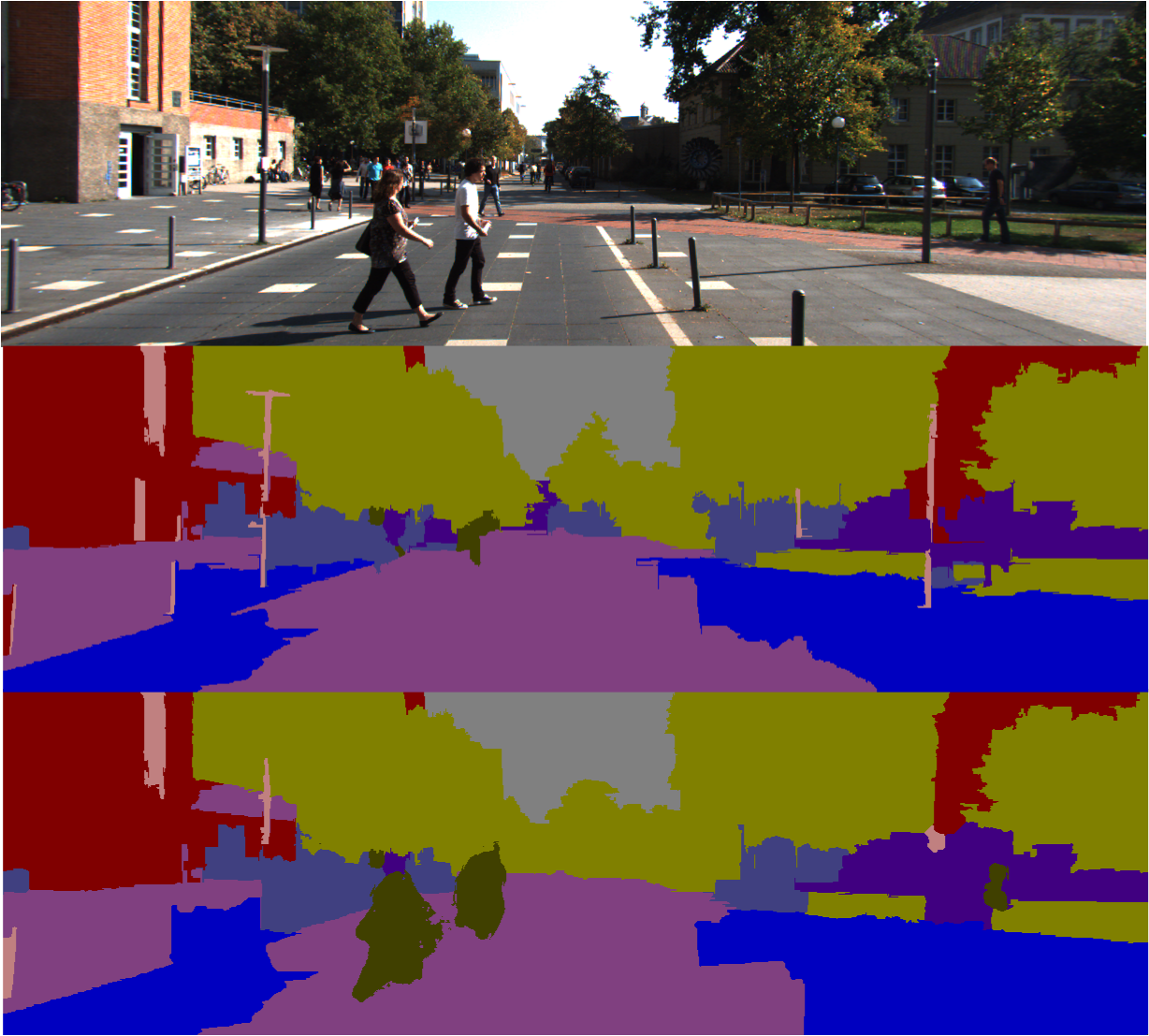}\\
\includegraphics[width=80mm]{misc/h}\\

\caption{
In the figure we do a comparative evaluation between the results of Full-C and our method.The original image(1) is taken from the KITTI dataset. The output of the Full-C(2) is depicted which shows a wrong labelling of the pedestrian pixels in the image. The Results of the proposed method(3) depict the improvement in the semantic segmentation.
}
\label{fig:ALE_123}
\end{center}

\end{figure}

 \textbf{Qualitative evaluation :} We show our results in comparison to Ground truth, in semantic segmentation with FULL-C and in motion to GEO-M Fig 2. In Sequence 1, FULL-C isn't able to segment cars as a whole and miss out on several patches while GEO-M fails in the case due to degeneracy in motion. Sequence 3, has patches on the front car due to failure of disparity computation, also the car's window is wrongly classified by ALE. These things are corrected by our method. The motion consistency helps in removing the window patch while in Sequence 1 the degeneracy is handled by our motion estimator independent of such geometric constraints. The joint formulation captures the motion and improves the semantics of the image. We also show an improvement in the segmentation where the disparity computation has failed as show in the Sequence 1. We take a specific example in Fig \ref{fig:ALE_123} showing Pedestrians, the image is smoothed out by the fully connected dense CRF leading to the wrong labelling while our method is able to correctly segment the whole pedestrian. This again reiterates the use of motion correlation for a better labelling. In all the above cases, we can see the effectiveness of our algorithm in handling the motion to generate a dynamic semantic model of the scene. We tried using motion cues as a feature in the object class unary. This couldn’t be used as a discriminative feature for an object class, as objects can have different motions which can not be learned through textonboost. We have implemented the Fully connected CRF module as it was showing substantial improvement in results for specific classes like pedestrian compared to a superpixel clique based CRF model.

 \textbf{Quantitative evaluation :} We quantitavely compare our approach against the other state-of-the-art image segmentation approaches, including pairwise CRF semantic segmentation approach with super-pixel based higher orders (AHCRF), Fully connected CRF (Full-C) and joint motion-object CRF with superpixel-clique consistency (JAHCRF). The quantitative evaluation of the object class segmentation of our joint optimization method with respect to other approaches is summarized in Table {\ref{tab:table_ale}. Evaluation is performed by cross verifying each classified pixel with the Ground truth .We choose the average intersection/union as the evaluation measure for both the image segmentation and the motion segmentation.It is defined as $TP/(TP+FP+FN)$, where TP represents the true positive ,FP the false positive and FN as the false negative. We observe an increase in performance for most of the classes in each of these measurements, mainly the object classes car and person have shown substantial improvements in accuracy. This is attributed to the fact that motion can be associated with specific classes and the pairwise connections in the motion domain respect the continuity in optical flow, while in the image domain,the connections between neighbouring pixels might violate the occlusion boundaries.
 \begin{table}
 \begin{center}
    \begin{tabular}{ | c | c  c  |}
    \hline
  Method & Moving & Stationary \\ \hline
     & & \\
         GEO-M \cite{NamdevKKJ12}     &\ \ \ \ 46.5  \ \ \ \ & \ \  49.8\ \\ \hline
             & & \\
    AHCRF+Motion & \ \ \ \ 60.2  \ \ \ \  & \ \ 75.8 \ \\ \hline
          & & \\
    OURS-M & \ \ \ \ \bf 73.5  \ \ \ \  & \ \ \bf 82.4 \ \\ \hline
    \end{tabular}
   \caption{Quantitative analysis of motion segmentation for the Kitti dataset. We have compared our results(OURS-M) with geometric based motion segmentation(GEO-M) and joint optimization with superpixel based clique and motion estimate(AHCRF+Motion) }
   \label{tab:table_motion}
\end{center}
  \end{table}
 \begin{table*}[t!]
\begin{center}
    \begin{tabular}{ | c | c  c  c  c  c  c  c  c  c  c|}
    \hline
  Method & \begin{sideways} Pavement \end{sideways} & \begin{sideways} Road \end{sideways} & \begin{sideways} Sky \end{sideways} & \begin{sideways} Car \end{sideways} & \begin{sideways} Building \end{sideways} &  \begin{sideways} Vegetation \end{sideways} & \begin{sideways} Poles \end{sideways} & \begin{sideways} Pedestrian \end{sideways} & \begin{sideways} Fence \end{sideways} & \begin{sideways} Signage \end{sideways}  \\ \hline
      Super-pixel higher orders & & & & & & & & & &\\
           \bf (AHCRF)\cite{LadickyRKT09} & \ 74.2 \ & 93.2 \ & 95.2 \ & 77.9 \ & 94.2 \ & 84.5 \ & \bf 31.7 \ & 32.2 \ & 50.4 \ & 10.3 \\ \hline
     Fully-Connected CRF        & & & & & & & & & &\\
    \bf (Full-C)\cite{Krahenbuhl_Koltun_2011} & 73.5 & 93.4 & 95.3 & 77.4 & \bf 94.7 & 84.6 & 27.3 & 31.3 & 50.2 & 11.2\\ \hline
     AHCRF + Motion            & & & & & & & & & &\\
    \bf (JAHCRF) & \bf 73.7 & 93.7 & 95.2 & 81.2 & 94.5 & 84.7 & 31.5 & 37.2 & \bf 50.6 & 10.2\\ \hline
       Full-C + Motion         & & & & & & & & & &\\
    \bf (OURS-O)  & 73.6 & \bf 93.8 & \bf 95.3 & \bf 85.2 & 94.5 & \bf 84.7 & 27.1 & \bf 39.2 & 50.4 & \bf 11.4\\ \hline

    \end{tabular}
   \caption{This table shows the image-based semantic Evaluation for all the sequences of the KITTI dataset.We compare our results with publicly available semantic segmentation .1) super-pixel Clique-based CRF(AHCRF) 2)Fully connected CRF (FULL-C) 3)Joint motion and object class segmentation using clique (AHCRF-Motion) 4) Our method for Semantic segmentation.The table shows a substantial improvements in the object class segmentation of the car and pedistrian.  }
   \label{tab:table_ale}
\end{center}
\end{table*}

 The evaluation of motion segmentation is summarized in Table {\ref{tab:table_motion}.We compute our motion accuracy evaluation similar to the object class segmentation. We compare our results with geometric-based motion segmentation(GEO-M) \cite{NamdevKKJ12}, and joint labelling of motion and the superpixel based image segmentation results(AHCRF+Motion). We observe a increase in the accuracy of the motion segmentation. This increase in efficiency in the results is due to the fact that, we have incorporated the label and motion correlation. The fact that the possibility of a moving wall and moving tree is less compared to a moving car or a moving person,has been exploited.The results of motion segmentation have shown substantial improvement over (AHCRF+Motion) can be attributed to the robust pairwise potetials of the dense formulation.    
\begin{figure*}[t!]
\begin{tabular}{ c c c}

\begin{sideways} \bf \ \ \ Sequence 4 \end{sideways} & \includegraphics[width=82mm]{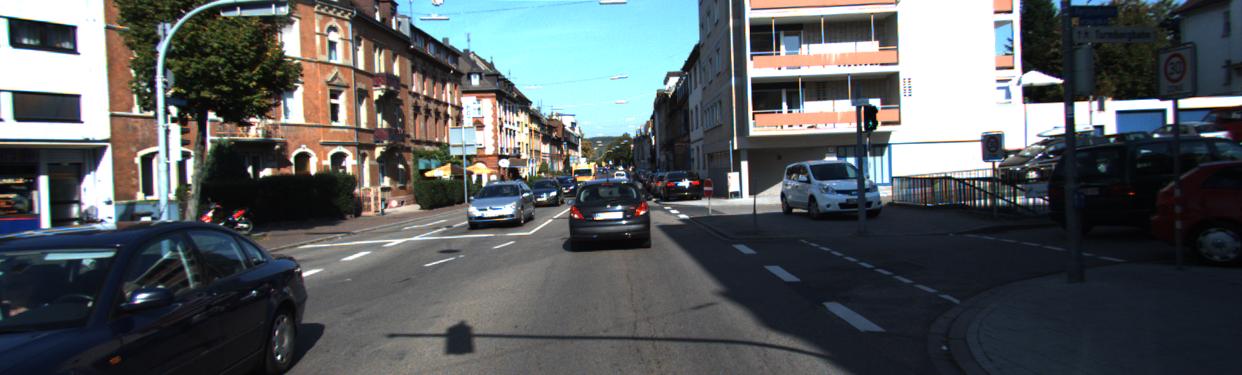} & \includegraphics[width=82mm]{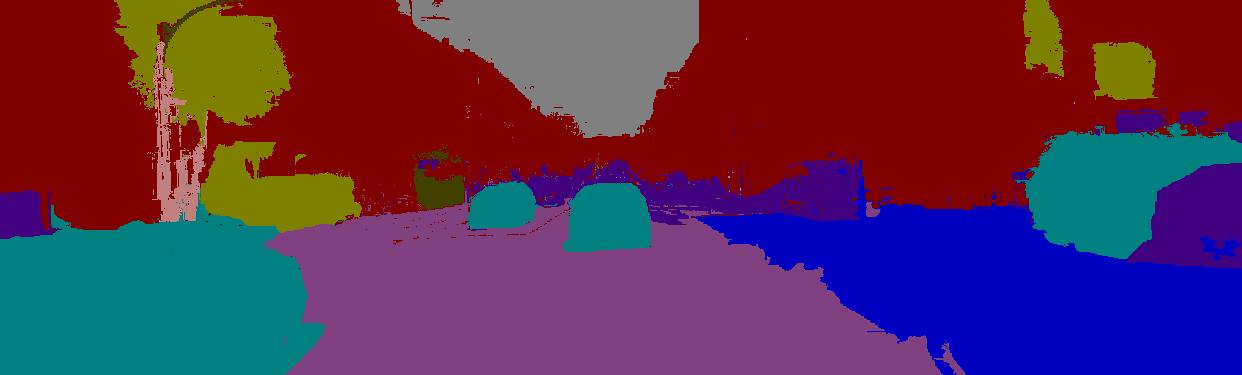} \\
 \begin{sideways} \bf \ \ \ Sequence 5 \end{sideways} & \includegraphics[width=82mm]{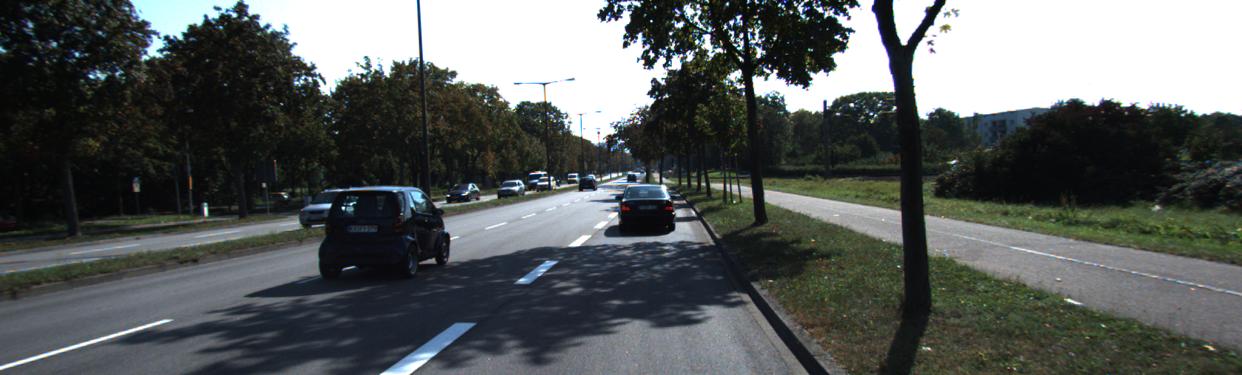} &  \includegraphics[width=82mm]{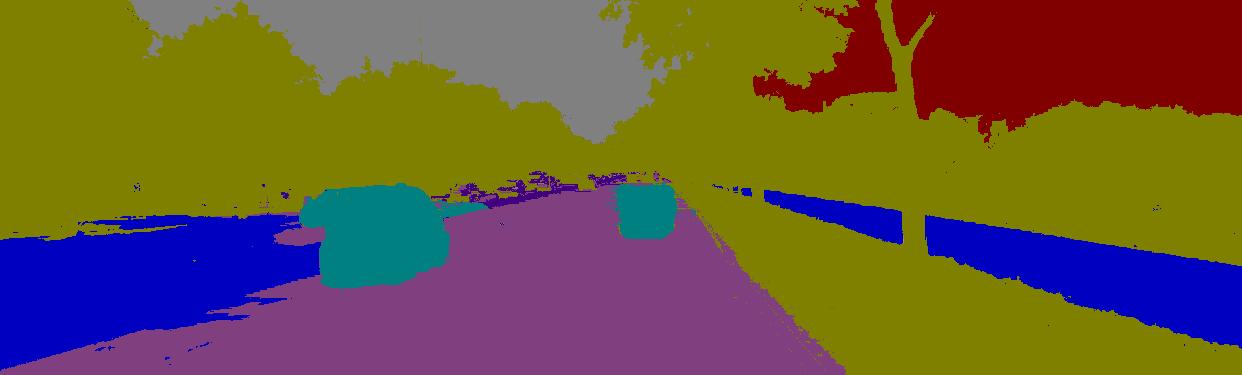} \\
\begin{sideways} \bf \ \ \ Sequence 6 \end{sideways} & \includegraphics[width=82mm]{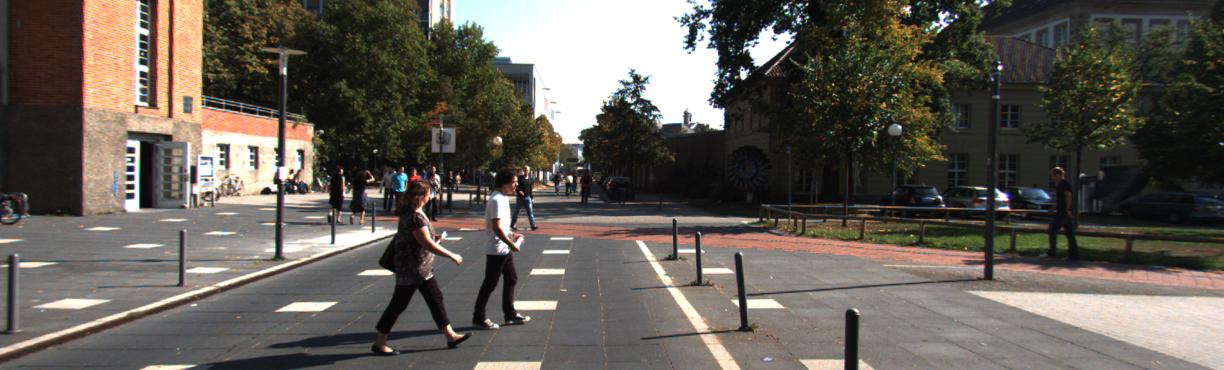} & \includegraphics[width=82mm]{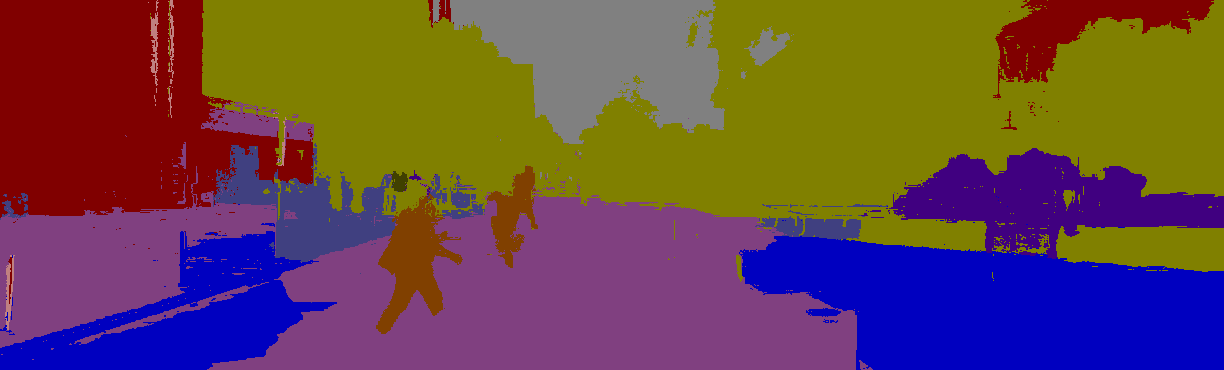} \\
 & \multicolumn{2}{c}{\includegraphics[width=168mm]{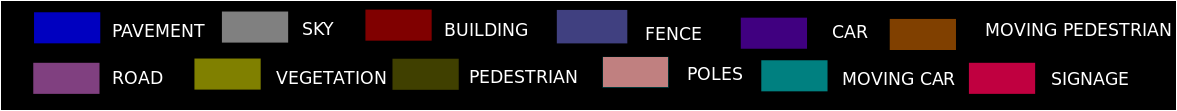}}
\end{tabular}

\caption{
The pixel wise result of our method on the KITTI test dataset . Note, that we are able to segment degenerate motions in the sequences. We show our results on varing scenarios i.e in a urban setting(Sequence 4) , highway setting(Sequence 5) and in case of moving pedistrian(Sequence 6). We achieve state-of-the art results for motion segmentation using our joint formulation.best viewed in color }
\label{fig:check}
\end{figure*}
\\

\section{conclusion and Future work}
\label{sec:conclusion}
In this paper ,we have proposed a joint approach simultaneously to predict the motion and object class labels for pixels and regions in a given image. The experiments suggest that combining information from motion and objects at region and pixel-levels helps semantic image segmentation .Further evaluations also show that per-pixel motion segmentation is important in achieveing higher accuracy in the motion segmentation results.In order to encourage future work and new algorithms in the area we are going to make the motion segmentation dataset of the KITTI tracking dataset available .

In the future work, We intend to extend the method by segmenting objects with different motion and segment each object as a different class.We also plan to achieve the GPU implementation for the proposed algorithm and generalize the current approach for dynamic scene understanding.We will continue expanding the annotations and the data in the KITTI tracking dataset.

\section{Acknowledgments}
This research was supported by the Department of Science and Technology grant SR/S3/EECE/0114/2010.





%

\bibliographystyle{abbrv}
\bibliography{sigproc}  
%
%

\end{document}